\documentclass{article}

\usepackage{PRIMEarxiv}

\usepackage[utf8]{inputenc} 
\usepackage[T1]{fontenc}    
\usepackage[hidelinks]{hyperref}       
\usepackage{url}            
\usepackage{booktabs}       
\usepackage{amsfonts}       
\usepackage{nicefrac}       
\usepackage{microtype}      
\usepackage{lipsum}
\usepackage{fancyhdr}       
\usepackage{graphicx}       
\graphicspath{{media/}}     

\usepackage{amsmath}
\usepackage{amsthm}
\usepackage{subcaption}
\usepackage{algorithm}
\usepackage{algorithmic}

\title{Graph Structural Residuals: A Learning Approach to Diagnosis}

\author{
  Jan Lukas Augustin \\
  Institute of Automation Technology \\
  Helmut Schmidt University, Hamburg, Germany \\
  \texttt{janlukas.augustin@hsu-hh.de} \\
   \And
  Oliver Niggemann \\
  Institute of Automation Technology \\
  Helmut Schmidt University, Hamburg, Germany \\
  \texttt{oliver.niggemann@hsu-hh.de} \\
}

\begin{document}
\maketitle

\begin{abstract}
Traditional model-based diagnosis relies on constructing explicit system models, a process that can be laborious and expertise-demanding. In this paper, we propose a novel framework that combines concepts of model-based diagnosis with deep graph structure learning. This data-driven approach leverages data to learn the system’s underlying structure and provide dynamic observations, represented by two distinct graph adjacency matrices. Our work facilitates a seamless integration of graph structure learning with model-based diagnosis by making three main contributions: redefining the constructs of system representation, observations, and faults; introducing two distinct versions of a self-supervised graph structure learning model architecture; and demonstrating the potential of our data-driven diagnostic method through experiments on a system of coupled oscillators.
\end{abstract}

\section{Introduction}

Diagnosing faults in complex systems has become a cornerstone challenge in artificial intelligence and engineering. Historically, model-based diagnosis (MBD) has been the primary methodology, with two main lines of research championing distinct approaches. Consistency-based diagnosis, deeply rooted in the realms of logic and optimization, offers a structured approach emphasizing system representation, observations, and health states. In contrast, the fault detection and isolation (FDI) community, grounded in engineering and control theory disciplines, leans towards continuous models that dynamically represent the states of systems. While undeniably robust and powerful, these approaches come with a clear challenge: the manual construction of system models, which is not only effort-intensive but also demands specialized expertise.

Parallel to the challenges faced in the traditional MBD arena, deep learning has emerged as a transformative force in learning complex representations directly from data. While Convolutional Neural Networks (CNNs) \cite{lecuncnn} and Long Short-Term Memory networks (LSTMs) \cite{hochreiter1997long} have redefined time-series data analysis, Graph Neural Networks (GNNs) have shown unparalleled aptitude in learning from structured data. Notably, GNNs have demonstrated exceptional performance in static datasets such as citation networks, but their potential in dynamic systems like cyber-physical systems remains under-explored \cite{Niggemann2023}.

Our work bridges this divide. Driven by the need for more flexible and adaptive diagnostic tools, we introduce a novel, learning-centric approach to model-based diagnosis that integrates deep graph structure learning with traditional diagnostic paradigms. At the heart of our approach is the representation of systems as spatiotemporal graphs. We present two novel models: a reference model that discerns a static representation of a system's normative behavior, and an observation model that produces dynamic graph structures based on current inputs. The primary novelty lies in identifying faults as discrepancies between these learned graph structures, marking a paradigm shift from traditional component health state assessments to the identification of faulty edges.

Our goal revolves around a pivotal question: can we redefine the principles of system representation, observations, and faults in the context of graph structural learning, and leverage this redefinition to craft a data-driven diagnostic mechanism? Pursuant to this, our research revolves around the following queries:
\begin{enumerate}
\item Can we learn a system's inherent structure, represented as the adjacency matrix of a graph, purely from observational data?
\item Is it feasible to construct a model that dynamically crafts observations, encapsulated as adjacency matrices, contingent on windows of multivariate time-series inputs?
\item Can the discrepancies between a learned static structure and dynamic observations serve as a basis to detect faults?
\end{enumerate}

Reflecting on these research questions, we make the ensuing contributions:
\begin{enumerate}
\item We present a novel redefinition of system representation, observations, and faults, enabling an elegant confluence of graph structure learning with the tenets of MBD. This foundational perspective drives our data-driven diagnostic approach.
\item We elucidate two distinctive graph structure learning models in a self-supervised way. One is trained to provide a static reference adjacency matrix encapsulating a normative system representation, while the other dynamically generates observation adjacencies based on current system inputs.
\item Through empirical analyses on a dataset of coupled oscillators,
we showcase the efficacy of our diagnostic method.
\end{enumerate}

In doing so, we aim to harmonize model-based diagnosis and deep learning, using the strengths of both to enhance the diagnostic capabilities. Our method \textbf{G}raph \textbf{S}tructural \textbf{R}esiduals (GSR) seeks to contribute to the research field by providing an innovative, data-driven approach to diagnosis. By learning directly from data, we expect to alleviate the need for manual model creation, paving the way for more automated and scalable diagnostic systems.

\section{Related Work}

Model-based diagnosis research is an active field that has attracted interest from various disciplines, leading to multiple distinct research tracks. Notably, consistency-based diagnosis and fault detection and isolation can both be regarded as prominent tracks of MBD \cite{trave2019bridge}. Despite their common interest in diagnosing systems, they have evolved independently, originating from different foundational backgrounds.

\textbf{Consistency-based diagnosis} is rooted in the fields of logic, combinatorial optimization, search, and complexity analyses. The focus lies in finding minimal sets of components that, when assumed faulty, make the system's behavior consistent with the observations \cite{de1987diagnosing, reiter1987theory}.
MBD has been studied in numerous domains including automotive \cite{struss2003model}, space \cite{williams1996model}, robotics \cite{khalastchi2013sensor}, software \cite{abreu2011simultaneous} and cyber-physical systems \cite{diedrich2022residual}.
Classic approaches include GDE \cite{de1987diagnosing}, CDA* \cite{williams2007conflict} or  SDE \cite{stern2012exploring}.
Components play an instrumental role in this model; their behavior, and the interplay among them, forms the cornerstone of diagnostic inferences.

On the other hand, the \textbf{fault detection and isolation} is deeply rooted in engineering disciplines, particularly control theory and statistical decision-making \cite{blanke2006diagnosis, de1987diagnosing}. An essential tool in FDI is the generation and use of residuals, which are computed as the difference between measured and expected system behaviors \cite{gertler1998fault}. Structural analysis is a prevalent approach in FDI, providing insights into system behaviors based on the interconnections between system components without necessarily diving into detailed quantitative models \cite{ducstegor2006structural, frisk2019structural}. It allows to check the diagnosability of a system and ensure optimal sensor placement \cite{krysander2008sensor}.

While traditional MBD emphasizes system models and relations of components, another emergent field seeks to harness the structure inherent in data itself: the realm of \textbf{graph neural networks} (GNNs) \cite{wu2020comprehensive} in which graphs serve as powerful data representation structures in various domains, ranging from social networks to molecular architectures. To capture the intricate relationships within graph-structured data, Graph Neural Networks (GNNs) have emerged as the forefront technology \cite{bronstein2021geometric}. GNNs operate on the principle that a node's state is influenced by its neighbors, effectively modeling complex graph patterns \cite{scarselli2008graph, bruna2013spectral}. One notable variant, Graph Convolution Networks (GCNs), aggregate the representations of a node's immediate neighbors to form its feature representation \cite{Kipf2016-vy, duvenaud2015convolutional}. Such techniques have paved the way for breakthroughs in diverse applications, from traffic prediction \cite{Yu2017-td, Chen2019-md} and recommendation systems \cite{Wu2022-af} to multi-relational data tasks \cite{Schlichtkrull2018-ae}.

While GNNs have exhibited robust performance on predefined graph structures, there exists an intriguing avenue of research focused on learning or refining the graph structure from data. Termed as \textbf{graph structure learning} (GSL), this paradigm addresses scenarios where the graph topology is either not provided a priori or can be optimized using available data \cite{zhu2021survey}. GSL strategies typically revolve around adjustable adjacency matrices, representing adjacency with learnable parameters, and optimizing it in conjunction with GNNs for tasks like node classification \cite{franceschi2019learning, yu2021graph, jin2020graph, wang2021graph, chen2020iterative}.
While deep GSL approaches have shown promise, they often depend on a supervised scenario. In the absence of labels we use a self-supervised approach inspired by SLAPS \cite{Fatemi2021-vy}.


\section{Preliminaries and Notation}
\label{sec:preliminariesandnotation}

In this section, we lay the foundational concepts required to understand our approach, along with the accompanying notation. Our work takes a graph-centric view on fault diagnosis by considering both the normative system representation and the observation to be outputs of learnable deep learning models. Consequently, faults are calculated as deviations between reference and observation graphs.

\subsection{Architecture}
\begin{figure}[ht]
    \centering
    \includegraphics[width=0.65\linewidth]{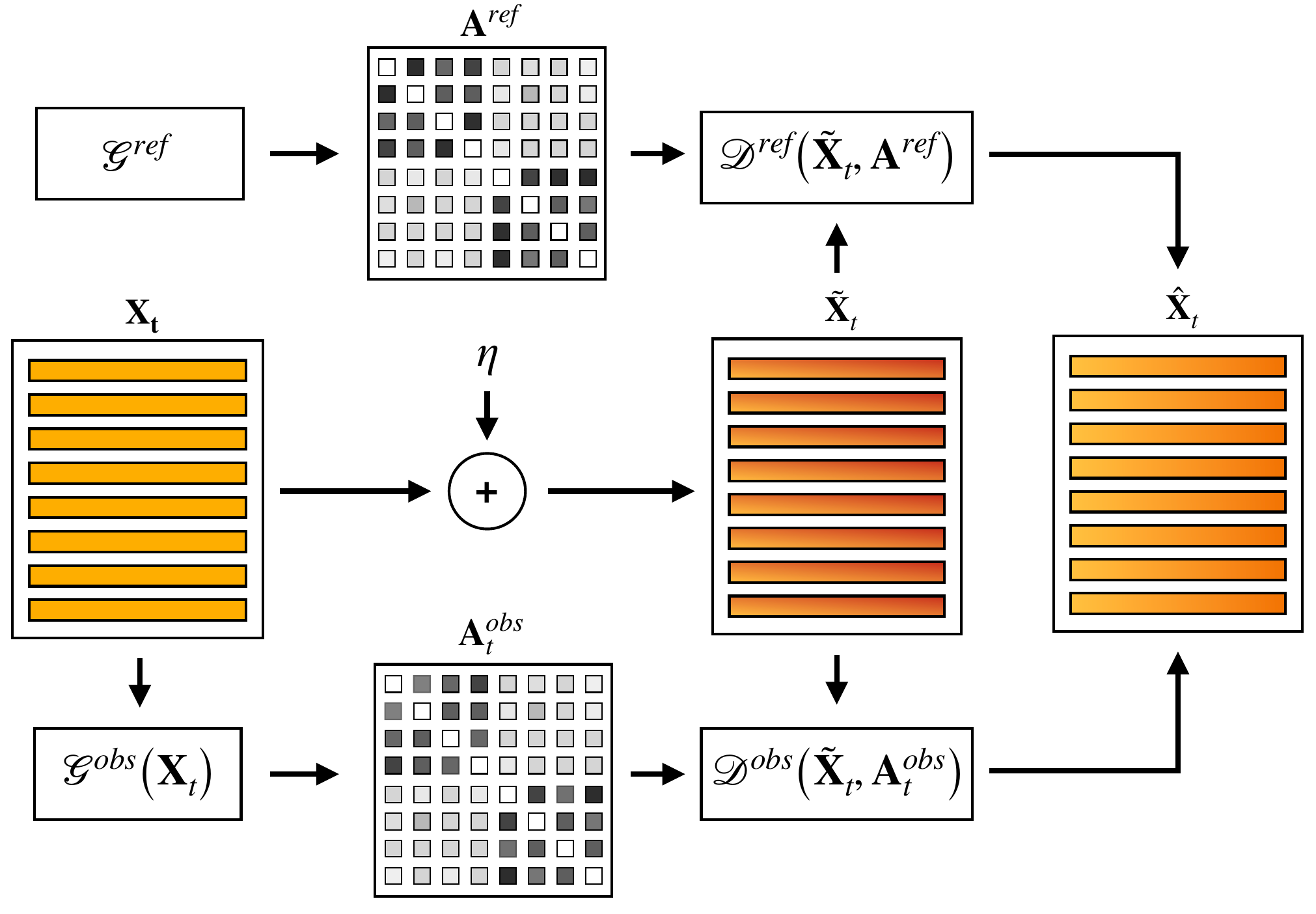}
    \caption{Overview of the GSR architecture during training. Given a noisy version $\mathbf{\tilde{X}}_t$ of input features $\mathbf{X}_t$ that are generated from a system with an unknown underlying structure, denoising modules $\mathcal{D}^{ref}$ and $\mathcal{D}^{obs}$ are trained to reconstruct the original features using adjacencies $\mathbf{A}^{ref}$ and $\mathbf{A}^{obs}_t$ generated by graph generators $\mathcal{G}^{ref}$ or $\mathcal{G}^{obs}$, respectively.}
    \label{fig:gsl4dtraining}
\end{figure}

While various graph structure learning algorithm may be used to generate the necessary adjacency matrices, in the abscence of labels we suggest a self-supervised approach. As shown in Figure \ref{fig:gsl4dtraining}, the two-stage model architecture consists of a graph generator module and a denoising module, which are trained end-to-end to minimize the reconstruction error. The reconstruction error serves as an indirect supervision signal to guide the graph generator to output weighted edges that resemble the connections of the underlying structure from which the data has been generated.

\subsection{Multivariate time series}

A multivariate time series is a collection of observations from multiple variables recorded over the same time intervals. Such data is often represented as a matrix, where each row corresponds to a variable and each column represents a time step. Given the length of the multivariate time series $T$, number of variables $N$, length of the time window $W$ and artificially generated noise $\eta$ we can define:

\paragraph{Definition 1}
\label{par:definition1}
(Multivariate Time Series Window).\\
A window of multivariate time series data starting at time \( t \) and ending at time \( t+W \) is represented by the matrix \( \mathbf{X}_{t:t+W} \) where \( \mathbf{X}_{t:t+W} \in \mathbb{R}^{N \times W} \). If \( \eta_{t:t+W} \) denotes noise in this window, a corrupted version is given by:
$$\mathbf{\tilde{X}}_{t:t+W} = \mathbf{X}_{t:t+W} + \eta_{t:t+W}$$

To improve readability, we refer to $\mathbf{X}_{t:t+W}$ as $\mathbf{X}_t$.

\subsection{Spatiotemporal graphs}

Spatiotemporal graphs introduce a temporal dimension into classical graph structures, enabling the representation of interactions and relationships that evolve over time \cite{rozemberczki2021pytorch}.
We take a discrete temporal snapshot view by considering time windows and introduce two types of graphs: Static spatiotemporal graphs that do not evolve over time are used to obtain the reference structure and dynamic spatiotemporal graphs that depend on the temporal signal provide the observation structure.
Given a graph $G$ where nodes $V$ represent variables and an adjacency matrix $\mathbf{A}$ where \( \mathbf{A} \in \mathbb{R}^{N \times N} \) depicts their relationships we define two types of spatiotemporal graphs.


\paragraph{Definition 2.1}
\label{par:definition21}
(Static Spatiotemporal Graph).\\
Let \( G = \{ V, \mathbf{A}, \mathbf{X}_t \} \) be a spatiotemporal graph where the strength and direction of relationships are quantified by its adjacency matrix \( \mathbf{A} \), and each element \( A^{ij} \) represents the weight of the edge between node \( i \) and node \( j \).

\begin{figure}[ht]
    \centering
    \includegraphics[width=0.7\linewidth]{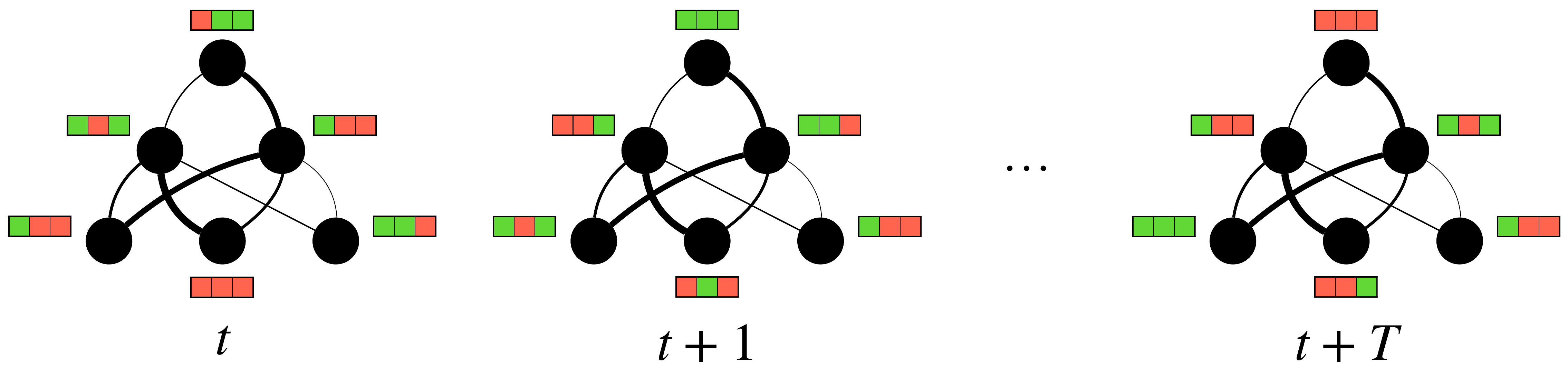}
    \caption{Static spatiotemporal graph}
    \label{fig:staticgraphvis}
\end{figure}

\paragraph{Definition 2.2}
\label{par:definition22}
(Dynamic Spatiotemporal Graph).\\
Let \( G = \{ V, \mathbf{A}_t, \mathbf{X}_t \} \) be a spatiotemporal graph where the strength and direction of relationships are quantified by its time-dependent adjacency matrix \( \mathbf{A}_t \), and each element \( A{_t^{ij}} \) represents the weight of the edge between node \( i \) and node \( j \) for a time window starting at time \( t \) and ending at time \( t+W \).

\begin{figure}[ht]
    \centering
    \includegraphics[width=0.7\linewidth]{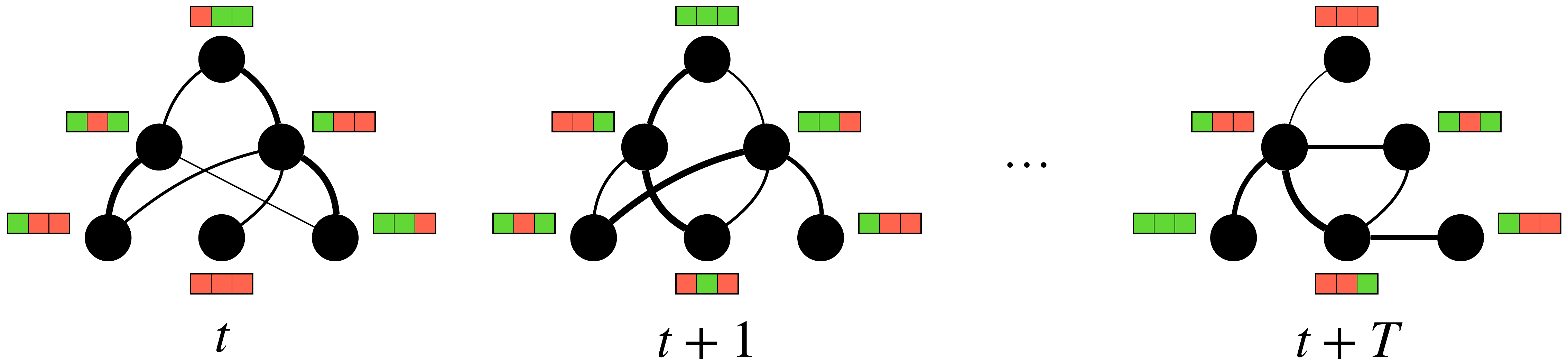}
    \caption{Dynamic spatiotemporal graph}
    \label{fig:dyngraphvis}
\end{figure}


\subsection{Graph Generator Module}
\label{sec:graphgenmodule}

The graph generator module is designed to be trained to generate the adjacency matrix that serves as an input of the denoising module. The generation happens depending on learnable parameters and - in the case of the observation graph generator - the multivariate time series. Given a graph generator function $\mathcal{G}(\cdot)$ and learnable parameters of the graph generator function $\theta_{\mathcal{G}}$ we can define:

\paragraph{Definition 3}
\label{par:definition3}
(Graph Generator Module).\\
For a time series data window \( \mathbf{X}_t \), the adjacency matrices \( \mathbf{A}^{ref} \) and \( \mathbf{A}^{obs}_t \) are outputted by the reference and observation graph generator modules, respectively, as:

\begin{align}
    \mathbf{A}^{ref} = \mathcal{G}^{ref}(\mathbf{\theta}_{\mathcal{G}^{ref}}) \\
    \mathbf{A}^{obs}_t = \mathcal{G}^{obs}(\mathbf{X}_t, \mathbf{\theta}_{\mathcal{G}^{obs}})
\end{align}

For better readability in the text, where applicable we also refer to $\mathbf{A}^{ref}$ and $\mathbf{A}^{obs}_t $ as $\mathbf{A}$.

\subsection{Denoising Module}
\label{sec:denoisingmodule}

The denoising module is built to be trained to eliminate or reduce the noise and reconstruct the original multivariate time series data given a noisy version of it and the adjacency matrix provided by the graph generator module. Given a denoising function $\mathcal{D}(\cdot, \cdot)$ and parameters of the denoising function \( \mathbf{\theta}_{\mathcal{D}} \) we can define:

\paragraph{Definition 4}
\label{par:definition4}
(Denoising Module).\\
For a noisy time series data window \( \mathbf{\tilde{X}}_t \), the denoised versions using the reference and observation models, \( \mathbf{\hat{X}}_t^{ref} \) and \( \mathbf{\hat{X}}_t^{obs} \), are calculated by $\mathcal{D}$ as follows:

\begin{align}
\mathbf{\hat{X}}_t^{ref} = \mathcal{D}^{ref}(\mathbf{\tilde{X}}_t, \mathbf{A}^{ref}, \mathbf{\theta}_{\mathcal{G}^{ref}}) \\
\mathbf{\hat{X}}_t^{obs} = \mathcal{D}^{obs}(\mathbf{\tilde{X}}_t, \mathbf{A}^{obs}_t, \mathbf{\theta}_{\mathcal{G}^{obs}})
\end{align}

\subsection{Graph Structural Residuals}
\label{sec:graphresiduals}

In systems characterized by graph structures, understanding the deviation or residual between the observed and reference relationships is pivotal. The residuals can provide insights into discrepancies, misalignments, or faults in the system. For the context of this work, we define a residual function that can be used after the model has been trained. It maps the reference and observation adjacency matrices to a residual matrix highlighting these deviations.

\paragraph{Definition 5}
\label{par:definition5}
(Graph Structural Residual).\\
The residual matrix, \( \mathbf{R}_t \), for a given time window starting at time \( t \) is a function of the difference between the reference and observed adjacency matrices \( \mathbf{A}^{ref} \) and \( \mathbf{A}^{obs}_t \). It can be defined using the function \( \mathcal{R}(\cdot, \cdot) \) as:

\begin{equation}
    \mathbf{R}_t = \mathcal{R}\bigl(\mathbf{A}^{ref}, \mathbf{A}^{obs}_t\bigr)
\label{eq:residuals}
\end{equation}

\begin{figure}[ht]
    \centering
    \includegraphics[width=0.5\linewidth]{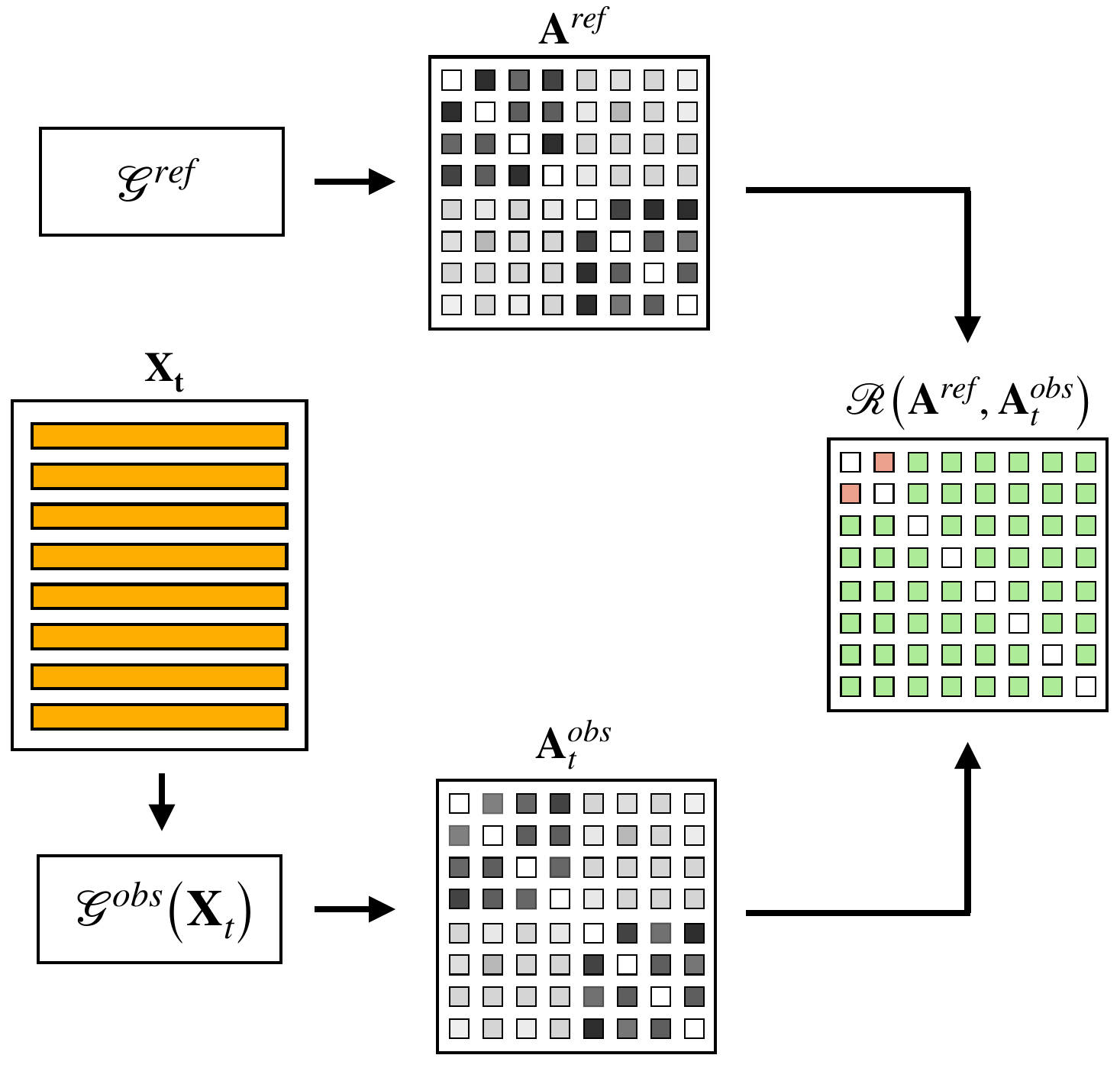}
    \caption{Overview of the GSR architecture generating residuals. Given input features $\mathbf{X}$ adjacencies $\mathbf{A}^{ref}$ and $\mathbf{A}^{obs}_t$ generated by graph generators $\mathcal{G}^{ref}$ or $\mathcal{G}^{obs}$, respectively. Graph structural residuals $\mathbf{R}_t$ are calculated as deviations between reference and observation to detect faulty edges.}
    \label{fig:gsl4doverview}
\end{figure}

For the simplest case, the residual function \( \mathcal{R} \) computes the absolute value of the difference between the reference and observed adjacency matrices. If a threshold \( \tau \) is defined, all entries in \( \mathbf{R}_t \) that are below \( \tau \) can be set to zero, denoting them as insignificant residuals. In essence, each entry \( R{_t^{ij}} \) in \( \mathbf{R}_t \) quantifies the deviation of the edge weight between node \( i \) and node \( j \) from the reference to the observed structure for the time window starting at \( t \).

\begin{equation}
    \mathbf{R}_t = \begin{cases} 
      |\mathbf{A}^{ref} - \mathbf{A}^{obs}_t| & \text{if } |\mathbf{A}^{ref} - \mathbf{A}^{obs}_t| \geq \tau \\
      0 & \text{otherwise}
   \end{cases}
\end{equation}

\section{Methodology}
\label{sec:methodology}

In the realms of fault diagnosis, CBD and FDI start out with manual modeling processes to represent system dynamics and behavior, allowing them to detect deviations and infer potential faults. In contrast, our proposed methodology hinges on leveraging learning-based mechanisms to capture relations which can be later employed for fault diagnosis. Conceptually, our methodology can be discerned into two core stages, akin to CBD and FDI, albeit with stark distinctions in the means of achieving the desired outcomes:

\paragraph{Model Construction:} Unlike traditional methods where this stage is manual, we employ a learning-based approach. Our model is trained to identify underlying system relationships.
\paragraph{Fault Diagnosis:} Post the construction (or training) of the model, we use it for generating graph structural residuals. These residuals, as previously introduced, provide insights into the discrepancies between the expected and observed system behaviors, thereby allowing us to pinpoint potential faults.

With this overarching view in mind, we delve deeper into the nuances of both stages in the following subsections.

\subsection{Model Construction}

By choosing a GNN for our denoising autoencoder module we make use of the smoothness assumption GNNs are built upon, which posits that connected nodes in a graph are likely to have similar features or labels \cite{zhu2020beyond}. The graph generator is incentivized to provide a graph with high homophily since the GNN requires information to come from relevant neighboring nodes to reconstruct the original features in the best possible way.

Given that inputs consist of time series data, we incorporate an additional inductive bias to guide the learning process by applying Temporal Convolutional Networks (TCNs) \cite{Bai2018-zs} in both the observation graph generator and the denoising autoencoder module. TCN ensures the model respects the causal structure of the time series data by using causal convolutions, makes nearby time steps more relevant than distant ones, utilizes dilated convolutions to increase the model's receptive field and includes residual connections to facilitate the flow of gradients.

\subsubsection{Graph Generator Modules}

Both generators implement a function $\mathcal{G}$ with parameters $\theta_{\mathcal{G}}$ which produces a matrix $\mathbf{A}$ as output. We apply an activation function to a preliminary adjacency $\mathbf{\tilde{A}}$ before symmetrizing and normalizing it to obtain the adjacency matrix $\mathbf{A}$ which is subsequently used by the GNN layers of the denoising module $\mathcal{D}$. For the reference graph generator we apply the exponential linear unit (ELU) to avoid gradient flow problems in case any edge becomes negative and for the observation graph generator we apply the sigmoid function. Note that while we could also generate directed graphs, i.e. non-symmetric adjacency matrices, in our experiments we decide to restrict our model to undirected graphs to stabilize training.
We consider the following two graph generators:

During training the \textbf{reference graph generator} $\mathcal{G}^{ref}$ directly optimizes the adjacency matrix fully parametrized by $\mathbf{\theta}_{\mathcal{G}^{ref}}$ while ignoring the input node features. Consequently, the generator provides a single static matrix $\mathbf{A}^{ref}$ that is optimized on all samples of the training set and represents the structure of the system during normal operation.

For the \textbf{observation graph generator} $\mathcal{G}^{obs}$ the parameters $\mathbf{\theta}_{\mathcal{G}^{obs}}$ correspond to the weights of a TCN and $\mathbf{\tilde{A}} = \mathcal{G}^{obs}(\mathbf{X},\mathbf{\theta}_{\mathcal{G}^{obs}}) = \mathtt{TCN}(\mathbf{X}_t) \cdot \mathtt{TCN}(\mathbf{X}_t)^T = \mathbf{X'}_t \cdot {\mathbf{X'}_t}^T$, where the function $\mathtt{TCN}: \mathbb{R}^{N\times W} \rightarrow \mathbb{R}^{N\times W}$ produces a matrix with updated node representations $\mathbf{X'}_t$. Dot product is applied as a similarity measure that maps $\mathbb{R}^{N\times W} \rightarrow \mathbb{R}^{N\times N}$. Note that $\mathtt{TCN}$ is applied to each univariate time series separately making $\mathcal{G}^{obs}$ permutation equivariant.

\subsubsection{Denoising Modules}

The task of the denoising module $\mathcal{D}$ is to take a noisy version $\mathbf{\tilde{X}}_t$ of the node features $\mathbf{X}_t$ and the generated adjacency matrix $\mathbf{A}$ as inputs and produce the updated denoised node features $\mathbf{\hat{X}}_t$ as output. We add independent Gaussian noise $\eta_t$ to $\mathbf{X}_t$ to obtain $\mathbf{\tilde{X}}_t = \mathbf{X}_t + \mathbf{\mathcal{N}}(\mu,\,\sigma^{2})^{N \times W}$.
We use the same GNN-based implementation for both the reference and the observation denoising modules. In the following we introduce our graph TCN layers.

\subsubsection{Graph TCN Layers}

GNNs use the graph structure and node features $\mathbf{x_i}$ to learn a representation vector of a node, $\mathbf{h_i}$. Representation of nodes are iteratively updated by aggregating representations of their neighbors. After k iterations of aggregation, a node’s representation captures information from its k-hop network neighborhood. 
We use a custom graph layer with similarities to the popular GIN layers \cite{Xu2018-hs}.
However, connections are weighted, we use a TCN instead of an MLP and only apply it to the representations of neighboring nodes, excluding self-loops. We use residual connections for all layers except the output layer to allow for a deeper network while also forcing the GNN to use $\mathbf{A}$.
Note that TCN parameters are shared within a layer, i.e. the same TCN is applied to each aggregated node representation of that layer.
In matrix form the hidden layers are defined by
\begin{equation}
    \mathbf{\tilde{X}}^{(k)}_t = \mathbf{I} \cdot \mathbf{\tilde{X}}^{(k-1)}_t  + \sigma \left( \mathtt{TCN} \left( \mathbf{A} \cdot \mathbf{\tilde{X}}^{(k-1)}_t \right) \right),
\end{equation}
while the output layer without residual connection is defined as
\begin{equation}
\mathbf{\hat{X}}^{(k)}_t = \mathtt{TCN} \left( \mathbf{A} \cdot \mathbf{\tilde{X}}^{(k-1)}_t \right)
\end{equation}
forcing the model to optimize $\mathbf{A}$.

\subsubsection{Training}
As outlined in algorithm \ref{alg:algorithmtrainref}, during training, for the reference model we minimize the loss function $\mathtt{L}$, in this case the mean-squared error loss:

\begin{align}
    \mathcal{L} = \mathtt{L}\bigl(\mathbf{X}_t, \mathcal{D}^{ref}(\mathbf{\tilde{X}}_t, \mathbf{A}^{ref}, \mathbf{\theta}_{\mathcal{D}^{ref}})\bigr) \\
    \mathcal{L} = \mathtt{L}\bigl(\mathbf{X}_t, \mathcal{D}^{obs}(\mathbf{\tilde{X}}_t, \mathbf{A}^{obs}_t, \mathbf{\theta}_{\mathcal{D}^{obs}})\bigr)
\end{align}
where
$\mathbf{A}^{ref} = \mathcal{G}^{ref}(\mathbf{\theta}_{\mathcal{G}^{ref}})$ and $\mathbf{A}^{obs}_t = \mathcal{G}^{obs}(\mathbf{X}_t, \mathbf{\theta}_{\mathcal{G}^{obs}})$.

\begin{algorithm}[tb]
\caption{Training models}
\label{alg:algorithmtrainref}
\textbf{Inputs}: Dataset $\mathcal{X}_{train}$, \texttt{EPOCHS}, \texttt{BATCH\_SIZE}\\
\textbf{Parameters}: $\mathbf{\theta}_{\mathcal{G}^{ref}}, \mathbf{\theta}_{\mathcal{D}^{ref}}, \mathbf{\theta}_{\mathcal{G}^{obs}}, \mathbf{\theta}_{\mathcal{D}^{obs}}$\\
\textbf{Output}: Trained modules $\mathcal{G}^{ref}, \mathcal{D}^{ref}$, $\mathcal{G}^{obs}, \mathcal{D}^{obs}$

\begin{algorithmic}[1] 
\STATE Initialize parameters and optimizer
\FOR {\texttt{model} in \texttt{\{reference,observation\}}}
    \FOR {\texttt{epoch} in \texttt{EPOCHS}}
        \STATE Shuffle $\mathcal{X}_{train}$
        \FOR{\texttt{batch} in $\mathcal{X}_{train}$ with size \texttt{BATCH\_SIZE}}
            \STATE Get corrupted version $\mathcal{\tilde{X}}_{batch}$ adding noise $\eta$
            \STATE Generate adjacency matrices $\mathbf{A}$
            \STATE Compute denoised version $\mathcal{\hat{X}}_{batch}$
            \STATE Compute reconstruction loss
            \STATE Backpropagate the loss and update parameters       
        \ENDFOR
    \ENDFOR
\ENDFOR
\end{algorithmic}
\end{algorithm}

\subsection{Fault Diagnosis}

As outlined in Algorithm \ref{alg:algorithmresiduals}, for robustness, we calculate the mean over a set of time windows $\mathcal{W} = \{w_1, w_2, \ldots, w_m\}$, where $m$ is the number of windows.
Each window $w_i$ is defined by the start step $s_i$ and fixed length $W$.
Given that the observation adjacency matrix for each time window $w_i$ is $\mathbf{A}^{obs}_i(\mathbf{X}_i)$, the average observation adjacency matrix $\mathbf{A}^{obs}_{avg}(\mathcal{W})$ over the set of time windows can be calculated as:
\begin{equation}
    \mathbf{A}^{obs}_{avg}(\mathcal{W}) = \frac{1}{m}\sum_{i=1}^{m}\mathbf{A}^{obs}_i(\mathbf{X}_i)
\label{eq:adjobswindows}
\end{equation}

We can then calculate the residual matrix $\mathbf{R}_t$ for a set of time windows analogously to Equation \ref{eq:residuals}:

\begin{equation}
    \mathbf{R}_t(\mathcal{W}) = \bigl| \mathbf{A}^{ref} - \mathbf{A}^{obs}_{avg}(\mathcal{W}) \bigr|
\end{equation}

\begin{algorithm}[tb]
\caption{Fault diagnosis using GSR for a set of windows}
\label{alg:algorithmresiduals}
\textbf{Inputs}: $\mathcal{W}$, $\mathcal{G}^{ref}, \mathcal{D}^{ref}$, $\mathcal{G}^{obs}, \mathcal{D}^{obs}$\\ 
\textbf{Output}: Residual matrix $\mathbf{R}_t$ for a set of time windows $\mathcal{W}$

\begin{algorithmic}[1]
\FOR{each time window in $\mathcal{W}$}
    \STATE Generate $\mathbf{A}^{ref}$ and $\mathbf{A}^{obs}_t$
\ENDFOR
\STATE Calculate $\mathbf{A}^{obs}_{avg}(\mathcal{W})$
\STATE Compute the residual matrix $\mathbf{R}_t(\mathcal{W})$
\end{algorithmic}
\end{algorithm}

\section{Experiments}
\label{sec:experiments}

\subsection{Dataset}
\label{sec:datasets}

We generate our synthetic dataset using simulations of phase-coupled oscillators \cite{Kuramoto1975-yk}. The Kuramoto model is a nonlinear system of phase-coupled oscillators that can exhibit a range of complicated dynamics based on the distribution of the oscillators’ internal frequencies and their coupling strengths. We use the common form for the Kuramoto model of $n$ oscillators given by the following differential equation:
$$
\frac{d \phi_i}{d t}=\omega_i+\sum_{j \neq i} k_{i j} \sin \left(\phi_i-\phi_j\right)
$$
with phases $\phi_i$, coupling constants $k_{ij}$, and intrinsic frequencies $\omega_i$.
We simulate eight phase-coupled oscillators in 1D with intrinsic frequencies $\omega_i \sim \mathbf{\mathcal{N}}(\mu,\,\sigma)^{n \times d}$ and initial phases $\phi_i^{t=1}$ uniformly sampled from $[0, 2\pi)$.
We create two subsystems of connected oscillators by coupling all of the first four as well as all of the last four oscillators.

We connect pairs of oscillators $v_i$ and $v_j$ by setting the corresponding coupling constant $k_{ij}=k_{ji}=k$. All other coupling constants are set to 0. The resulting matrix $\mathbf{K}$ describes the underlying structure of the system of coupled oscillators and can therefore be treated as the ground truth coupling matrix $\mathbf{C}$ when using our proposed method.



\subsubsection{Training and Validation}
The model is built to denoise multivariate time series with $W$ time steps. We run the simulation for $2W$ time steps so that with each run of the simulation we obtain a signal matrix $\mathbf{S} \in \mathbb{R}^{N\times 2W}$. Samples of the training set $\mathcal{X}_{train}$ comprise the first $W$ time steps whereas the validation set $\mathcal{X}_{val}$ includes the last $W$ time steps of the corresponding simulation runs.

Parameters of the Kuramoto model such as intrinsic frequencies and coupling constants are chosen in such a way that within the first W time steps, depending on the initial phases, the oscillators are generally not yet fully synchronized making the training task non-trivial. In contrast, validation data covers a time span in which the system is already largely synchronized. This makes the reconstruction even more dependent on a correct adjacency matrix $\mathbf{A}$.

\subsubsection{Diagnosis}
Diagnosis data is generated analogously to training and validation data. However, after $2W$ time steps we change the underlying structure by shuffling $\mathbf{C}$ to replace it with a perturbed coupling matrix $\mathbf{\hat{C}}$. The simulation continues for another $3W$ steps and the system starts to synchronize and reach a new stable state. These last $3W$ steps are considered for diagnosis.

\subsection{Experimental Setup}

We train two models as outlined in Algorithm \ref{alg:algorithmtrainref}. Baseline configurations include:

\begin{itemize}
    \item \textbf{True}, which assumes that the underlying structure is known. $\mathbf{A}^{true} = \mathbf{C}$ replaces the output of the graph generator,

    \item \textbf{Correlations}, which replaces the output of the graph generator with $\mathbf{A}^{corr}$, a correlation matrix calculated by averaging the Paerson correlation over all samples of the training set, and

    \item \textbf{Features}, which ignores the TCN in the dynamic observation graph generator to calculate similarity between nodes directly from the untransformed features.
    
\end{itemize}

\subsubsection{Adjacency error metric} 
We are interested in the agreement between the ground truth coupling matrix $\mathbf{C}$ and the inferred adjacency matrix $\mathbf{A}$.
Since self-loops are ruled out in order to force the denoising module $\mathcal{D}$ to make use of informative nodes, entries of the matrix diagonals are masked out and hence also not considered for the evaluation of $\mathbf{A}$.
To adjust for differences in scale, $\mathbf{A}$ is rescaled to range $[0,1]$ before calculating the adjacency error,
$e_{adj} = \mathtt{MAE}(\mathbf{A}, \mathbf{C})$ which indicates whether the model is learning the correct structure. 

\subsubsection{Reconstruction error metric} Even though minimizing the reconstruction error is merely used as a means for the model to learn the structure of the underlying system, it is still a metric worth tracking since it could be used to perform anomaly detection and trigger the diagnosis. Same as for the loss, the reconstruction error $e_{rec}$ is calculated as the mean squared error between the input signal $\mathbf{X}$ and the denoised output $\mathbf{\hat{X}}$,
$e_{rec} = \mathtt{MSE}(\mathbf{\hat{X}}, \mathbf{X})$. 

\subsubsection{Implementation details}
We implemented our model in PyTorch \cite{paszke2019pytorch} and used the Kuramoto package \cite{kuramoto} to run simulations utilizing five different seeds to account for randomness.
We trained the model for 50 epochs using the Adam \cite{kingma2014adam} optimizer with a learning rate of 0.001, and a batch size of 16. We identified suitable hyperparameters [layers: 2, dropout: 0.5] for the denoising autoencoder based on the reconstruction error on the validation set. For the TCN proven parameters were chosen based on the original publication \cite{Bai2018-zs}.
We ran all experiments on a n1-standard-8 instance with 8 vCPUs, 30 GB of memory and a single Tesla T4 GPU.

\subsection{Results}

\subsubsection{Learning the Reference Model}

\begin{figure}[ht]
     \centering
    \begin{subfigure}[b]{0.4\linewidth}
        \centering
        \includegraphics[width=0.7\linewidth]{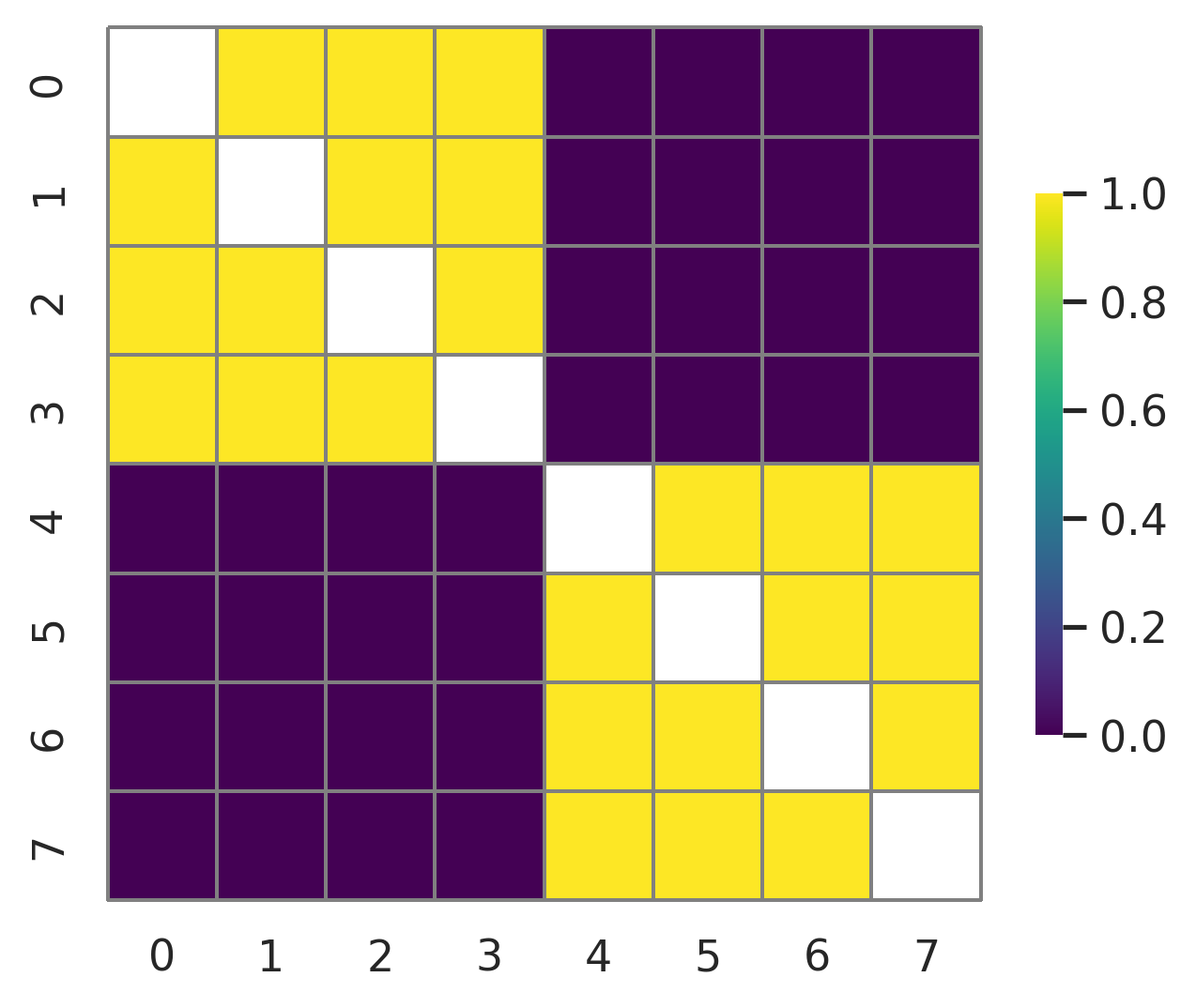}
        \caption{$\mathbf{C}$}
        \label{fig:groundtruthc}
    \end{subfigure}
    \hfill
    \begin{subfigure}[b]{0.4\linewidth}
        \centering
        \includegraphics[width=0.7\textwidth]{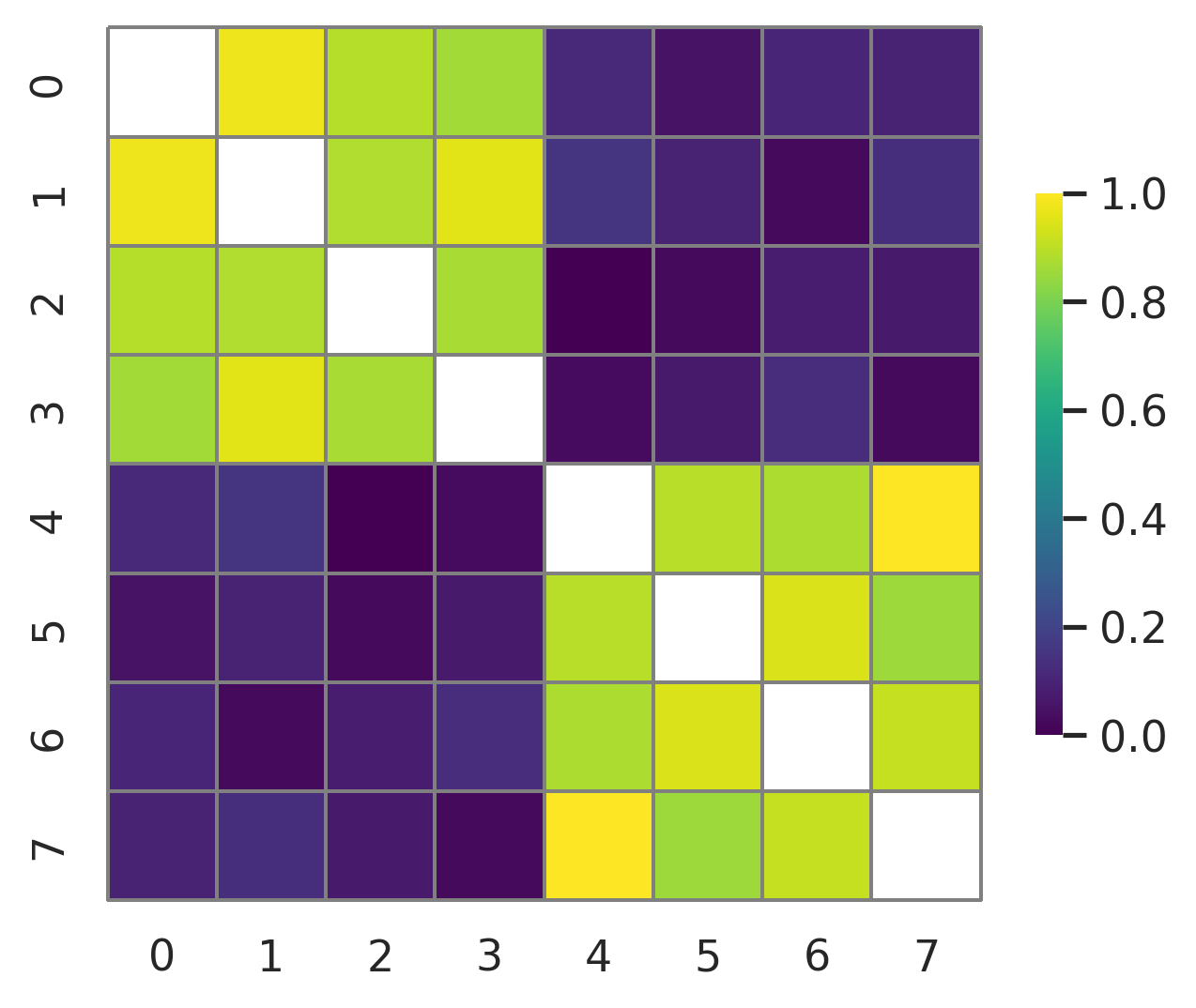}
        \caption{$\mathbf{A}^{ref}$}
        \label{fig:stat100test}
    \end{subfigure}
\caption{Comparison of coupling matrix $\mathbf{C}$ (a) and reference adjacency $\mathbf{A}^{ref}$ generated by $\mathcal{G}^{ref}$ (b)}
\end{figure}

Validation results given in Table \ref{tab:valresults} show that the $\mathcal{G}^{ref}$ generator improves upon the Correlation baseline in terms of adjacency error and yields results comparable to True and Correlation baselines with respect to reconstruction error. Figure \ref{fig:stat100test} shows the reference adjacency $\mathbf{A}^{ref}$ generated by $\mathcal{G}^{ref}$ alongside the ground truth $\mathbf{C}$. $\mathbf{A}^{ref}$ represents the average relations of nodes over all samples. Randomness introduced via different initial conditions of individual samples is reflected by variations in $\mathbf{A}^{ref}$ but $\mathbf{C}$ can be recovered by applying a threshold.

\begin{table}
\begin{center}
\caption{Validation results given as mean $\pm$ standard deviation for five different seeds.}
\label{tab:valresults}
\centering
\begin{tabular}{cccc}
Goal & Configuration & $e_{adj}$ & $e_{rec}$\\
\hline
\\[-6pt]
ref & True        & $0.000 \pm 0.000$ & $0.138 \pm 0.067$\\
ref & $\mathcal{G}^{ref}$      & $\mathbf{0.050 \pm 0.018}$ & $0.144 \pm 0.058$\\
ref & Correlation & $0.119 \pm 0.027$ & $0.139 \pm 0.057$\\
obs & $\mathcal{G}^{obs}$     & $0.260 \pm 0.009$ & $\mathbf{0.117 \pm 0.014}$\\
obs & Features    & $0.457 \pm 0.002$ & $0.240 \pm 0.034$\\
\hline
\end{tabular}
\end{center}
\end{table}

\subsubsection{Learning the Observation Model}

\begin{figure}[ht]
     \centering
     \begin{subfigure}[b]{0.4\linewidth}
         \centering
         \includegraphics[width=0.7\textwidth]{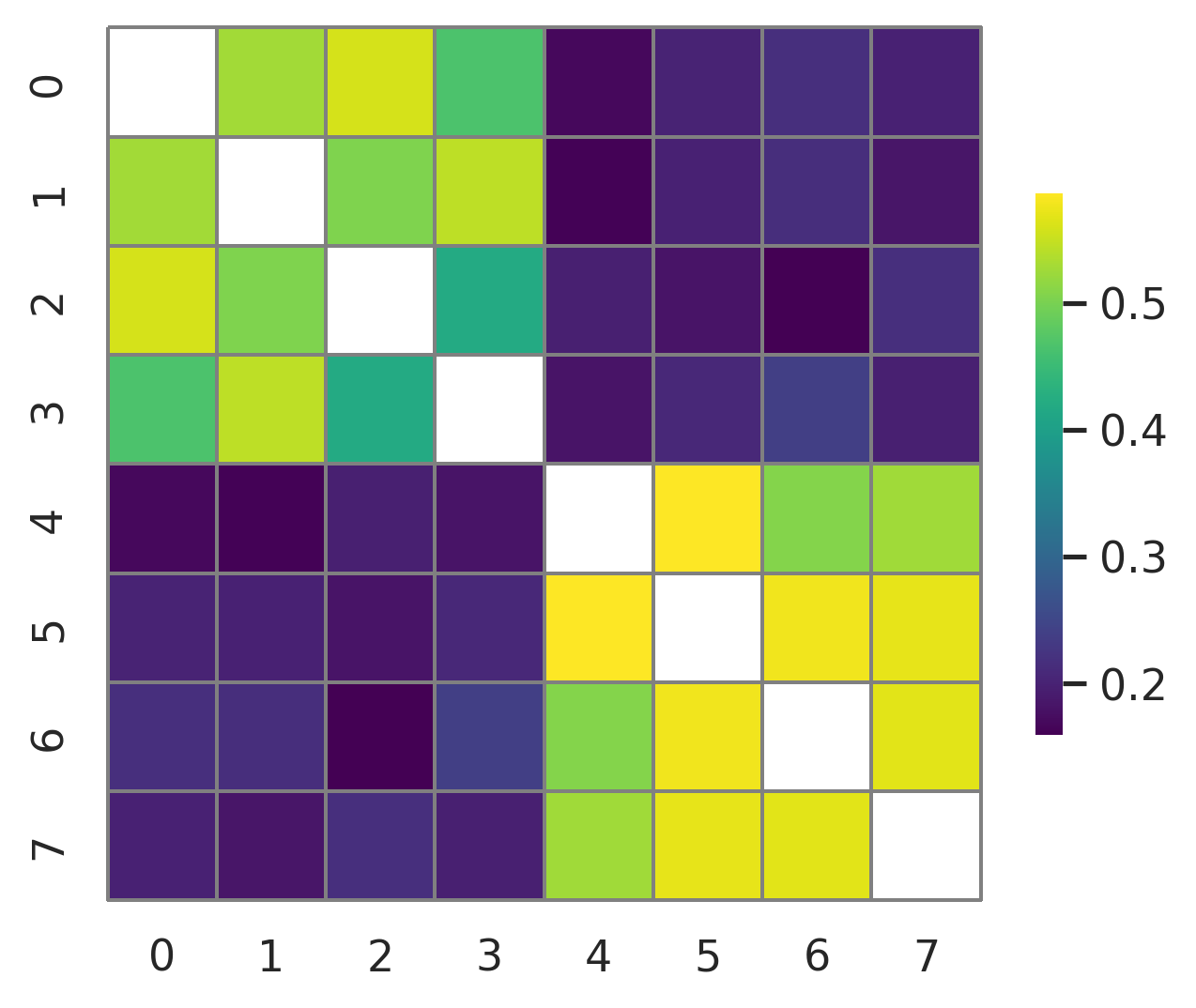}
     \end{subfigure}
    \hfill
     \begin{subfigure}[b]{0.4\linewidth}
         \centering
         \includegraphics[width=0.7\textwidth]{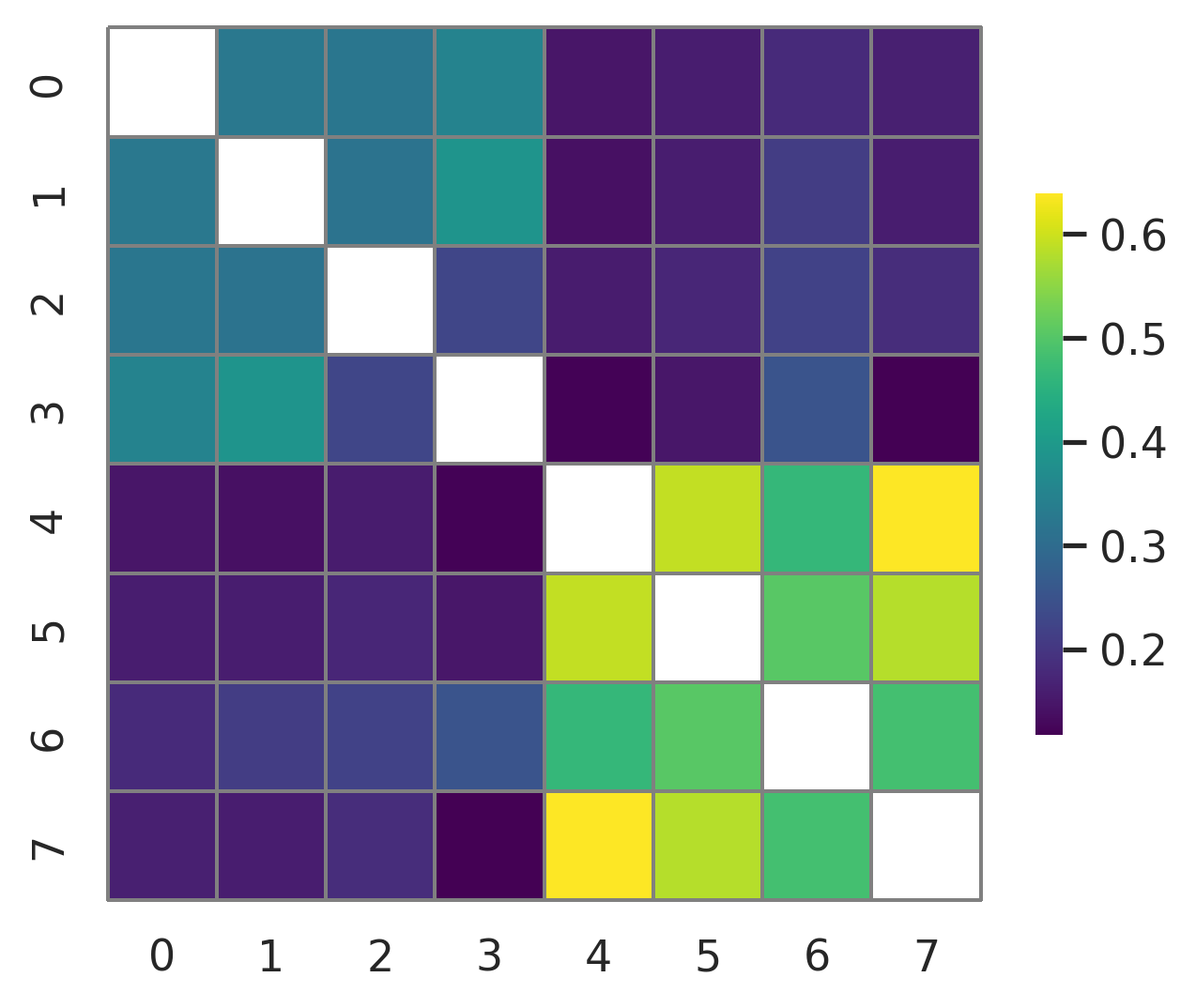}
     \end{subfigure}
        \caption{Examples of observation adjacency matrices $\mathbf{A}^{obs}$ generated by $\mathcal{G}^{obs}$ for two validation samples}
        \label{fig:valdynadjacencies}
\end{figure}

Validation results given in Table \ref{tab:valresults} show that the observation graph generator $\mathcal{G}^{obs}$ improves upon the Features baseline both in terms of adjacency error as well as with respect to reconstruction error.
Note that when comparing to $\mathcal{G}^{ref}$ we see that $\mathcal{G}^{obs}$ shows the best results with respect to reconstruction error but poorer adjacency error. This is expected since $\mathcal{G}^{ref}$ learns a set of parameters that best describe the system on average, while the dynamic $\mathcal{G}^{obs}$ can and will adjust the adjacency to the inputs. This behavior is illustrated in Figure \ref{fig:valdynadjacencies} which shows two observation adjacency matrices $\mathbf{A}^{obs}$ for two different input samples.

\subsubsection{Obtaining Graph Structural Residuals}

\subsubsection{Case 1 - Decoupling}
\begin{figure}[ht]
    \centering
    \begin{subfigure}[b]{0.5\linewidth}
        \centering
        \includegraphics[width=1.0\textwidth]{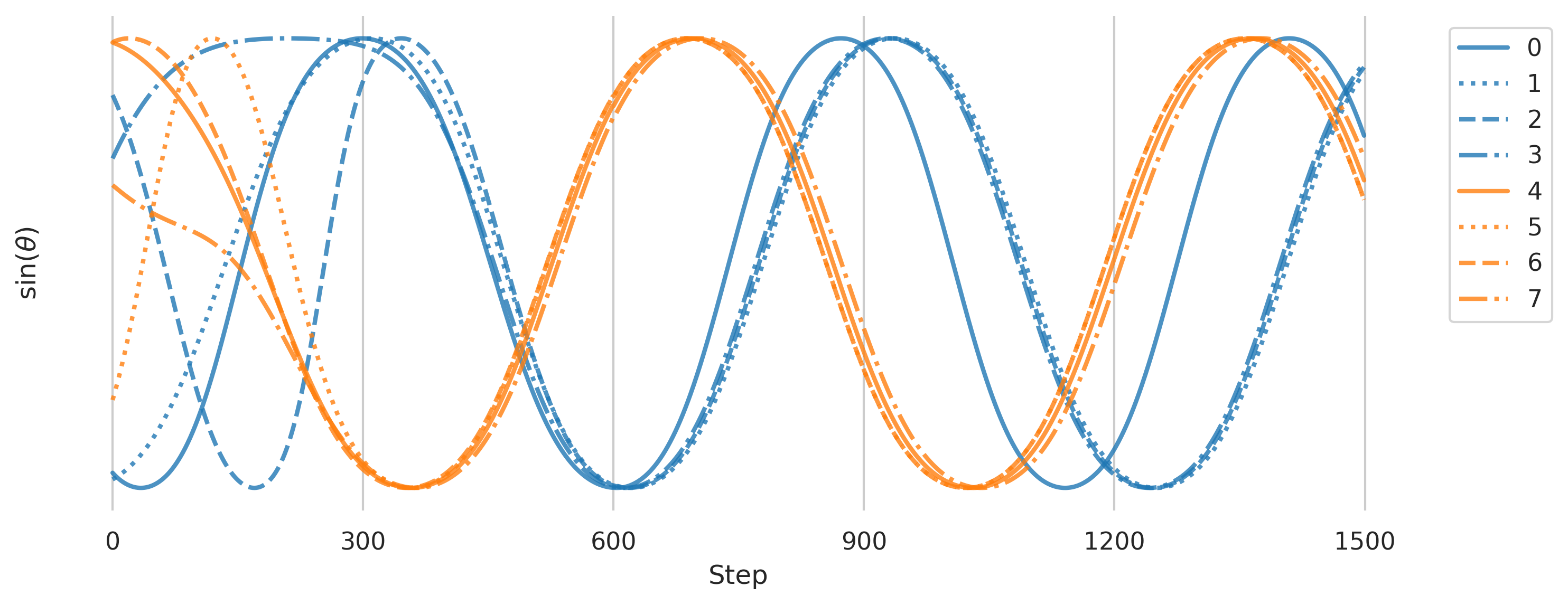}
        \caption{$\mathbf{S}$}
        \label{fig:decouplingsignals}
    \end{subfigure}
    \begin{subfigure}[b]{0.24\linewidth}
         \centering
         \includegraphics[width=0.99\textwidth]{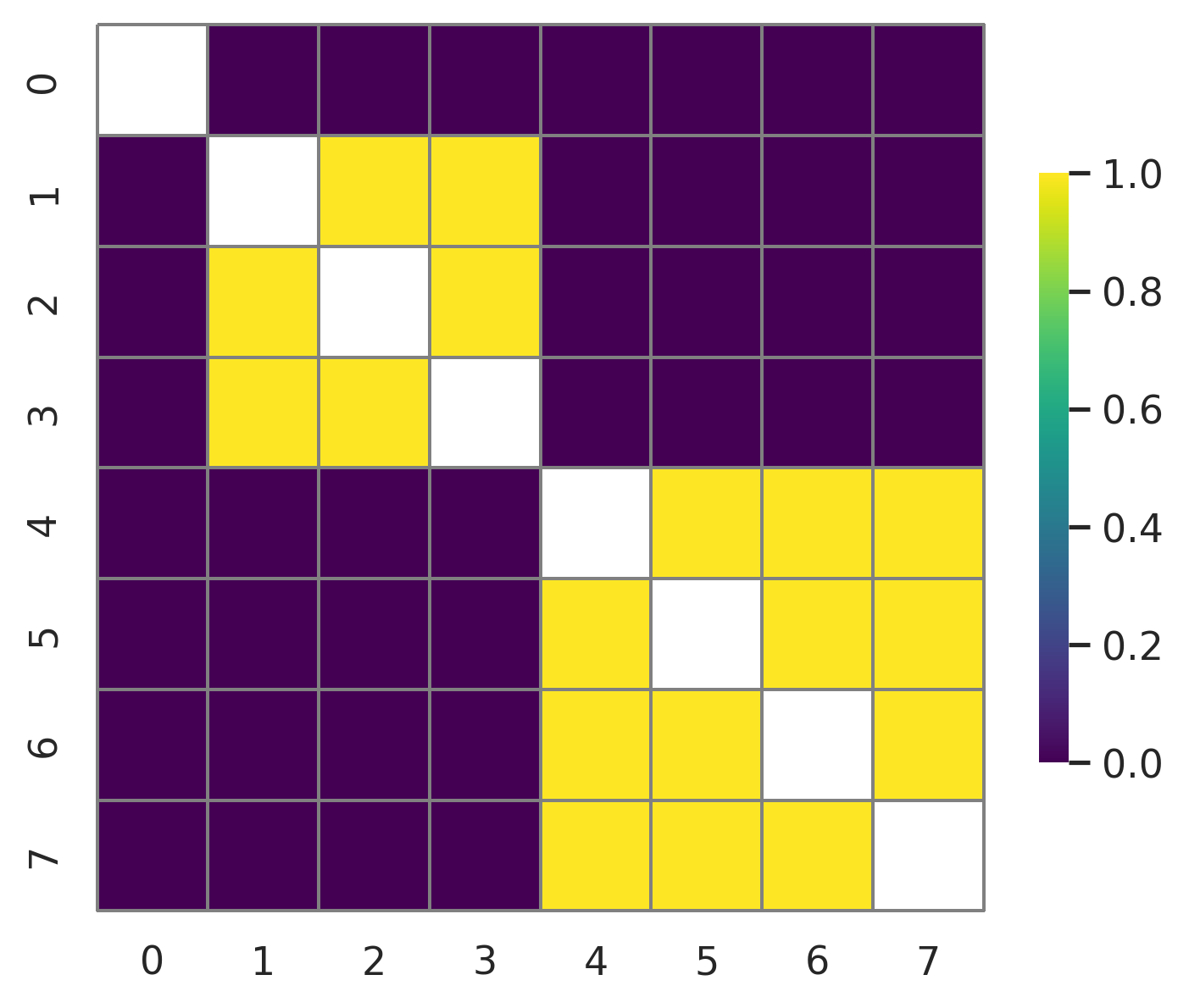}
         \caption{$\mathbf{\hat{C}}^{decoupling}$}
         \label{fig:cdecoupling}
     \end{subfigure}
    \hfill
    \begin{subfigure}[b]{0.24\linewidth}
         \centering
         \includegraphics[width=0.99\textwidth]{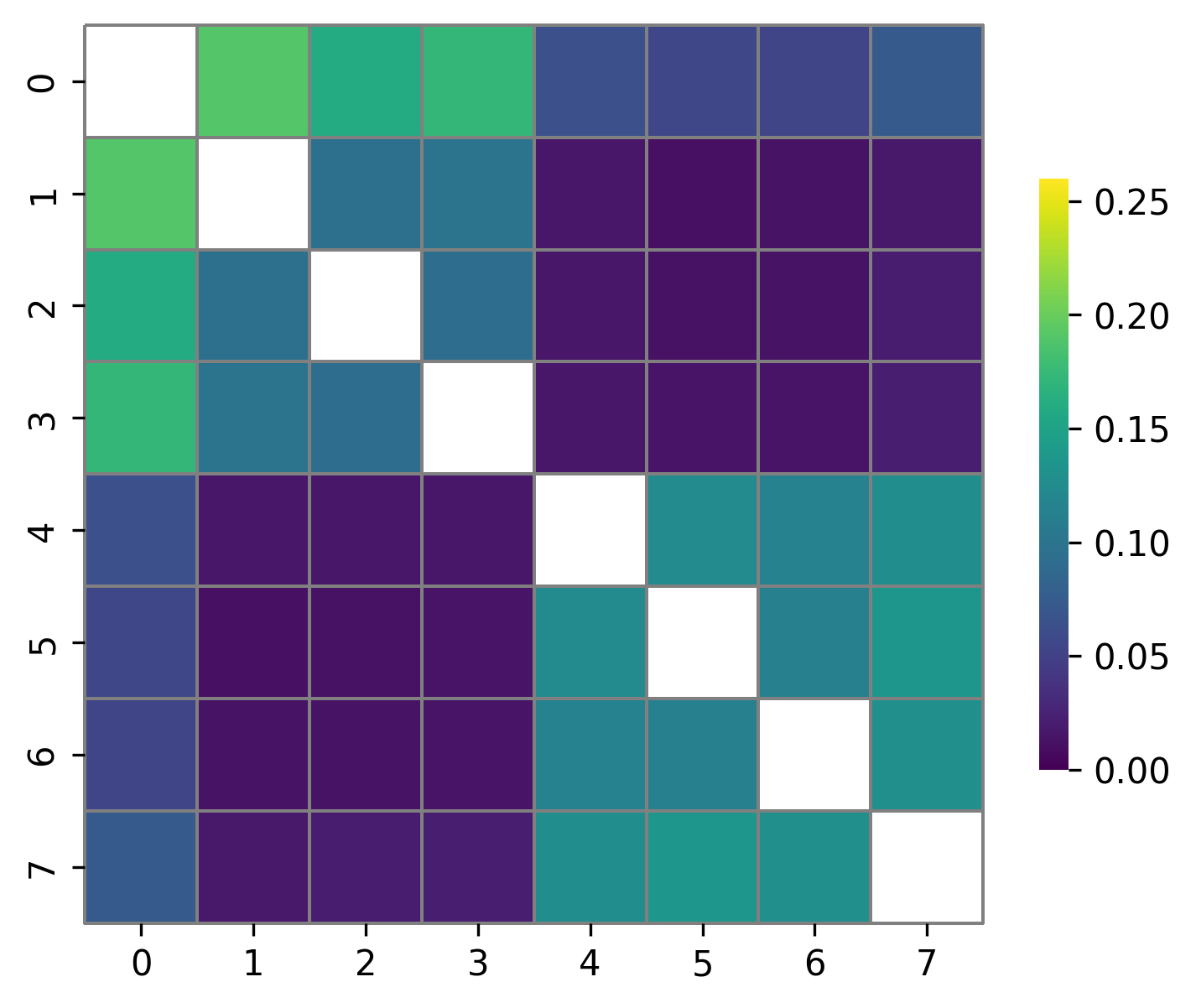}
         \caption{$\mathbf{R}^{decoupling}(\mathcal{W})$}
         \label{fig:aobsdecoupling}
    \end{subfigure}
        \caption{Case 1 - Decoupling. Signals (a), shuffled coupling matrix (b) and $\mathbf{R}^{decoupling}(\mathcal{W})$ for steps 600-1200 (c)}
        \label{fig:decouplingadjacenciesandsignals}
\end{figure}

For our first test case we consider a system in which a component is decoupled from its neighbors. We adjust $\mathbf{C}$ so that node 0 is decoupled from all other nodes in $\mathbf{\hat{C}}^{decoupling}$ (see Figure \ref{fig:cdecoupling}). Figure \ref{fig:decouplingsignals} shows the behavior of the system as node 0 desynchronizes following step 600, the point at which the underlying structure is changed. Inspection of $\mathbf{R}^{decoupling}(\mathcal{W})$ in Figure \ref{fig:aobsdecoupling} shows that connections involving node 0 which had been removed in $\mathbf{\hat{C}}^{decoupling}$ are highlighted.


\subsubsection{Case 2 - Swap}
\begin{figure}[ht]
    \centering
    \begin{subfigure}[b]{0.5\linewidth}
        \centering
        \includegraphics[width=1.0\textwidth]{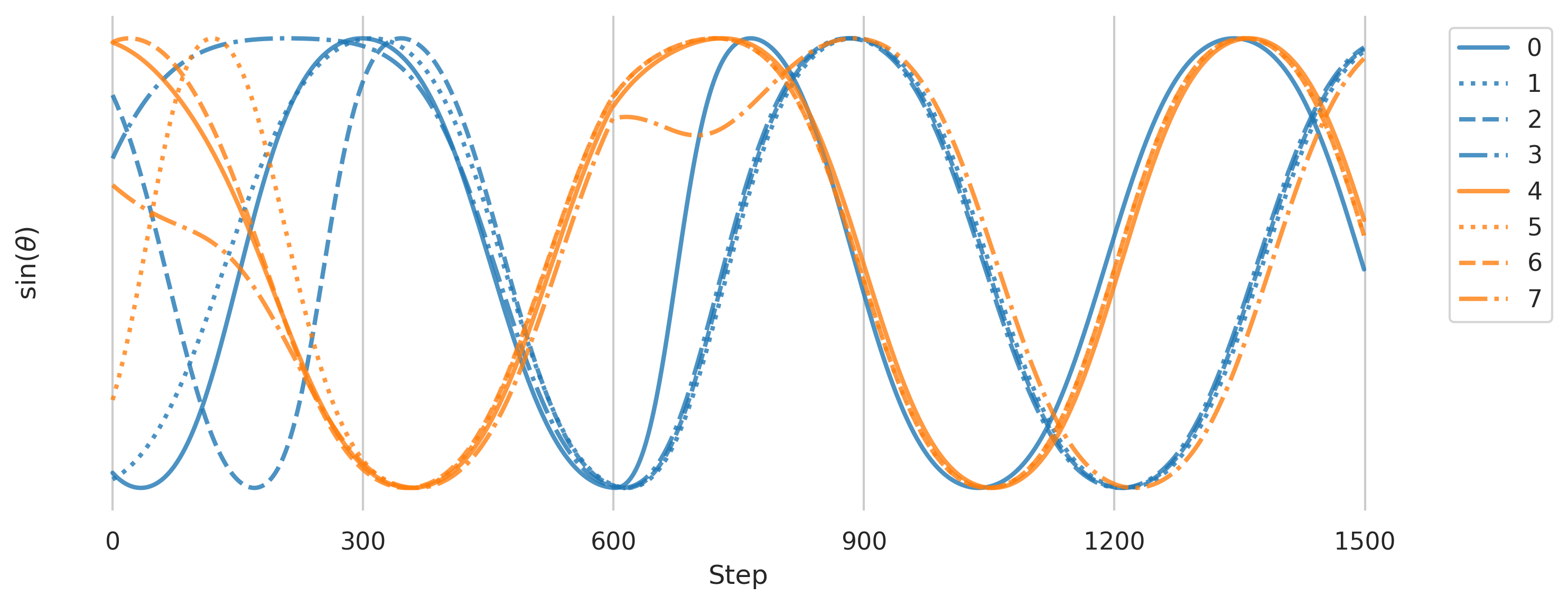}
        \caption{$\mathbf{S}$}
        \label{fig:swapsignals}
    \end{subfigure}
    \begin{subfigure}[b]{0.24\linewidth}
         \centering
         \includegraphics[width=0.99\textwidth]{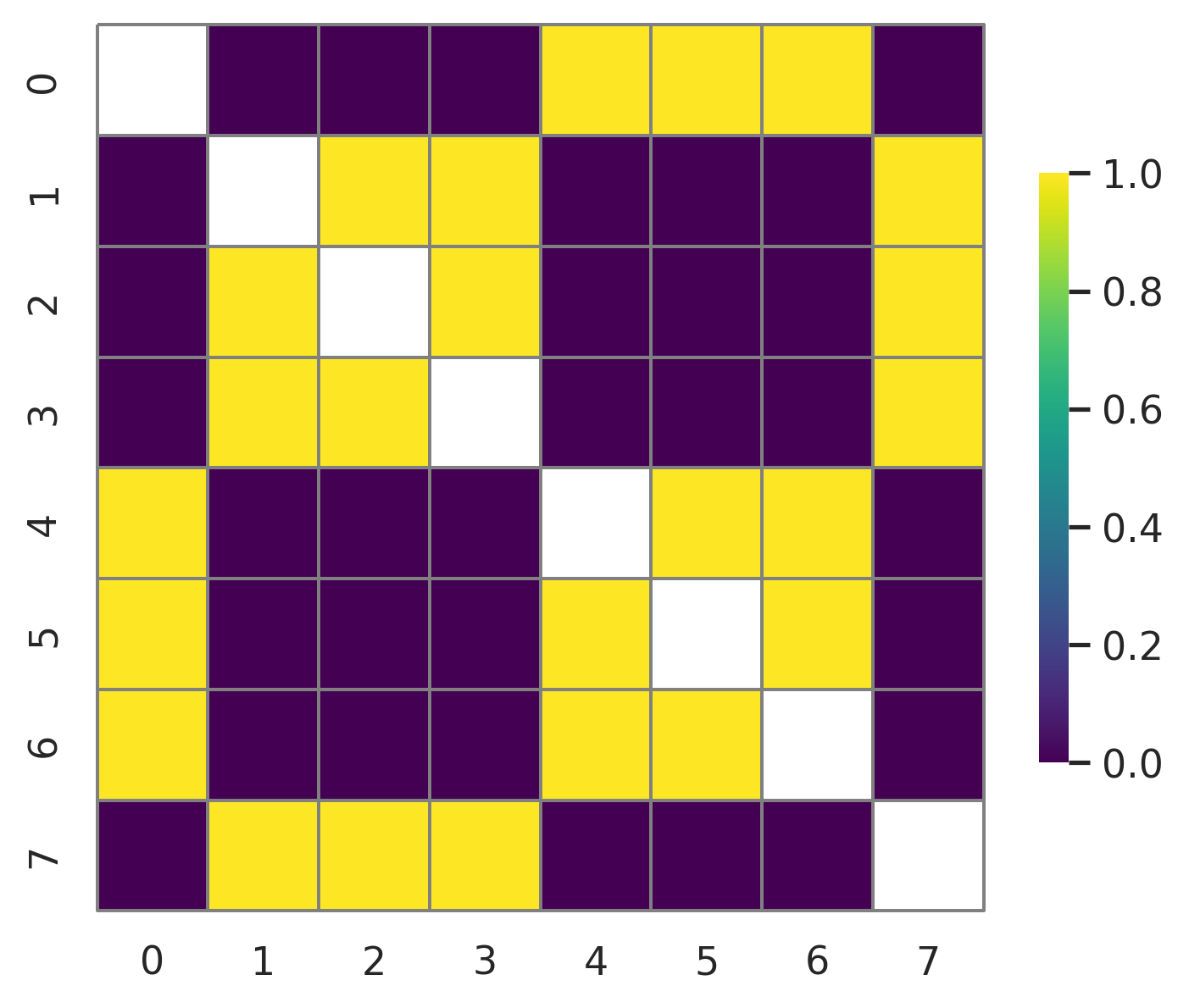}
         \caption{$\mathbf{\hat{C}}^{swap}$}
         \label{fig:cswap}
     \end{subfigure}
    \hfill
    \begin{subfigure}[b]{0.24\linewidth}
         \centering
         \includegraphics[width=0.99\textwidth]{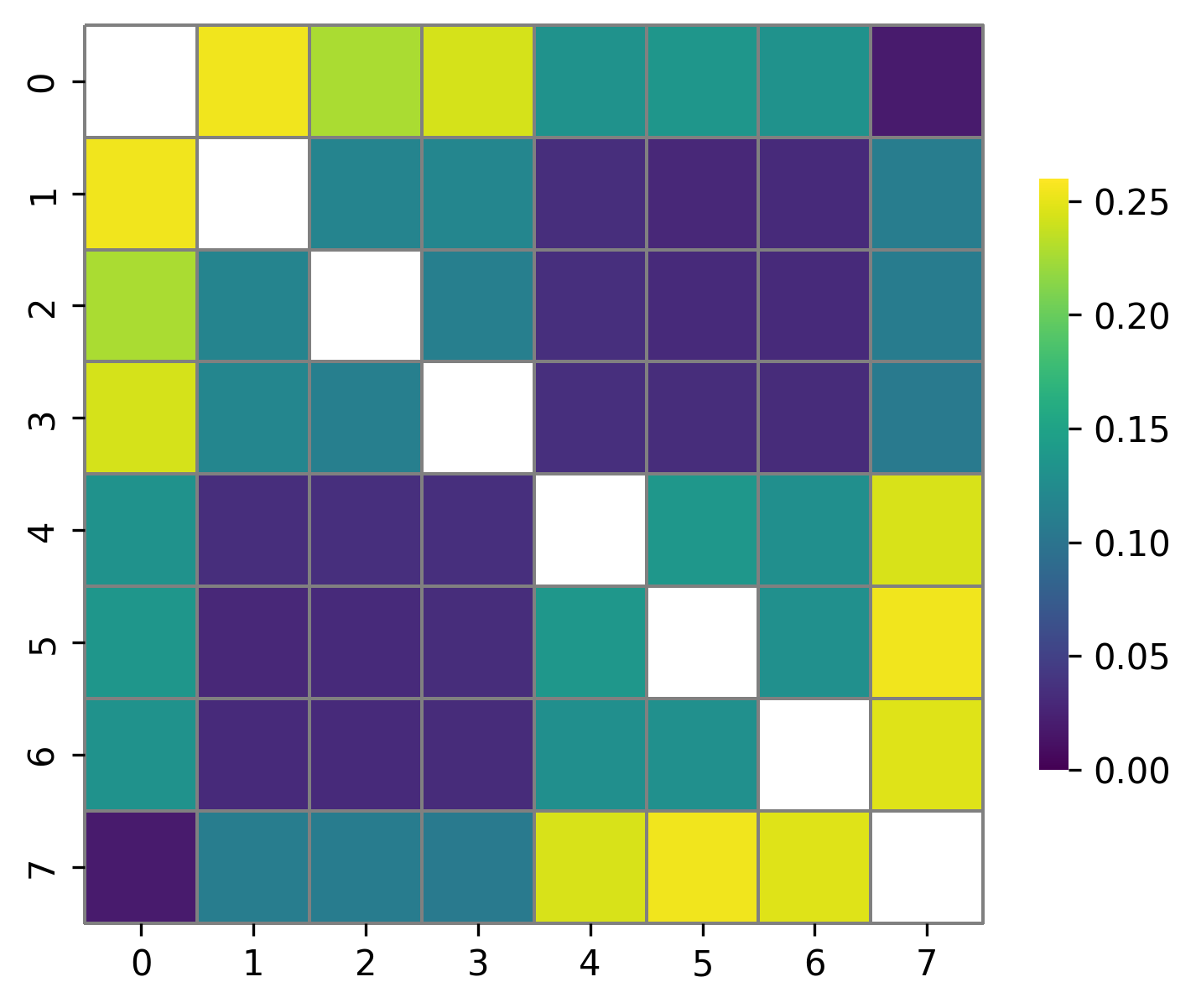}
         \caption{$\mathbf{R}^{swap}(\mathcal{W})$}
         \label{fig:aobsswap}
    \end{subfigure}
        \caption{Case 2 - Swap. Signals (a), shuffled coupling matrix (b) and $\mathbf{R}^{swap}(\mathcal{W})$ for steps 600-1200 (c)}
        \label{fig:swapadjacenciesandsignals}
\end{figure}

For our second test case we consider a system in which two components swap the subsystem to which they belong.
We adjust $\mathbf{C}$ so that node 0 and node 7 become coupled to the opposite set of oscillators in $\mathbf{\hat{C}}^{swap}$ (Figure \ref{fig:cswap}). Figure \ref{fig:swapsignals} shows the behavior of the system as nodes 0 and 7 synchronize to nodes 4-6 and nodes 1-3, respectively, following step 600.
Inspection of $\mathbf{R}^{swap}(\mathcal{W})$ in Figure \ref{fig:aobsswap} shows that indeed all connections involving node 0 or node 7 are highlighted.


\section{Conclusion}

In this paper, we presented a novel framework that merges traditional diagnosis concepts with deep graph structure learning for fault diagnosis. Key to our approach is the graph-based representation of the system, learned directly from data, negating manual system modeling. Our model dynamically processes multivariate time-series data, reflecting real-time system states, and introduces a new paradigm by directly inferring residuals from learned matrices. Nonetheless, our method's efficacy hinges on the data's quality and relevance. Further validation across diverse datasets and real-world situations is essential. Future research could refine the graph generation, integrate domain knowledge, or test the approach's versatility across domains.

\section*{Acknowledgments}
This research has been conducted as part of the project SmartShip which is funded by dtec.bw – Digitalization and Technology Research Center of the Bundeswehr. dtec.bw is funded by the European Union – NextGenerationEU.

\bibliographystyle{unsrt}  
\bibliography{references}

\begin{thebibliography}{10}

\bibitem{lecuncnn}
Y.~Lecun, L.~Bottou, Y.~Bengio, and P.~Haffner.
\newblock Gradient-based learning applied to document recognition.
\newblock {\em Proceedings of the IEEE}, 86(11):2278--2324, 1998.

\bibitem{hochreiter1997long}
Sepp Hochreiter and J{\"u}rgen Schmidhuber.
\newblock Long short-term memory.
\newblock {\em Neural computation}, 9(8):1735--1780, 1997.

\bibitem{Niggemann2023}
Oliver Niggemann, Bernd Zimmering, Henrik Steude, Jan~Lukas Augustin, Alexander
  Windmann, and Samim Multaheb.
\newblock {\em Machine Learning for Cyber-Physical Systems}, pages 415--446.
\newblock Springer Berlin Heidelberg, Berlin, Heidelberg, 2023.

\bibitem{trave2019bridge}
Louise Trav{\'e}-Massuy{\`e}s and Teresa Escobet.
\newblock Bridge: Matching model-based diagnosis from fdi and dx perspectives.
\newblock {\em Fault Diagnosis of Dynamic Systems: Quantitative and Qualitative
  Approaches}, pages 153--175, 2019.

\bibitem{de1987diagnosing}
Johan De~Kleer and Brian~C Williams.
\newblock Diagnosing multiple faults.
\newblock {\em Artificial intelligence}, 32(1):97--130, 1987.

\bibitem{reiter1987theory}
Raymond Reiter.
\newblock A theory of diagnosis from first principles.
\newblock {\em Artificial intelligence}, 32(1):57--95, 1987.

\bibitem{struss2003model}
Peter Struss and Chris Price.
\newblock Model-based systems in the automotive industry.
\newblock {\em AI magazine}, 24(4):17--17, 2003.

\bibitem{williams1996model}
Brian~C Williams, P~Pandurang Nayak, et~al.
\newblock A model-based approach to reactive self-configuring systems.
\newblock In {\em Proceedings of the national conference on artificial
  intelligence}, pages 971--978, 1996.

\bibitem{khalastchi2013sensor}
Eliahu Khalastchi, Meir Kalech, and Lior Rokach.
\newblock Sensor fault detection and diagnosis for autonomous systems.
\newblock In {\em Proceedings of the 2013 international conference on
  Autonomous agents and multi-agent systems}, pages 15--22, 2013.

\bibitem{abreu2011simultaneous}
Rui Abreu, Peter Zoeteweij, and Arjan~JC Van~Gemund.
\newblock Simultaneous debugging of software faults.
\newblock {\em Journal of Systems and Software}, 84(4):573--586, 2011.

\bibitem{diedrich2022residual}
Alexander Diedrich and Oliver Niggemann.
\newblock On residual-based diagnosis of physical systems.
\newblock {\em Engineering Applications of Artificial Intelligence},
  109:104636, 2022.

\bibitem{williams2007conflict}
Brian~C Williams and Robert~J Ragno.
\newblock Conflict-directed a* and its role in model-based embedded systems.
\newblock {\em Discrete Applied Mathematics}, 155(12):1562--1595, 2007.

\bibitem{stern2012exploring}
Roni Stern, Meir Kalech, Alexander Feldman, and Gregory Provan.
\newblock Exploring the duality in conflict-directed model-based diagnosis.
\newblock In {\em Proceedings of the AAAI Conference on Artificial
  Intelligence}, volume~26, pages 828--834, 2012.

\bibitem{blanke2006diagnosis}
Mogens Blanke, Michel Kinnaert, Jan Lunze, Marcel Staroswiecki, and Jochen
  Schr{\"o}der.
\newblock {\em Diagnosis and fault-tolerant control}, volume~2.
\newblock Springer, 2006.

\bibitem{gertler1998fault}
Janos Gertler.
\newblock {\em Fault detection and diagnosis in engineering systems}.
\newblock CRC press, 1998.

\bibitem{ducstegor2006structural}
Dilek D{\"u}{\c{s}}teg{\"o}r, Erik Frisk, Vincent Cocquempot, Mattias
  Krysander, and Marcel Staroswiecki.
\newblock Structural analysis of fault isolability in the damadics benchmark.
\newblock {\em Control Engineering Practice}, 14(6):597--608, 2006.

\bibitem{frisk2019structural}
Erik Frisk, Mattias Krysander, and Teresa Escobet.
\newblock Structural analysis.
\newblock {\em Fault Diagnosis of Dynamic Systems: Quantitative and Qualitative
  Approaches}, pages 43--68, 2019.

\bibitem{krysander2008sensor}
Mattias Krysander and Erik Frisk.
\newblock Sensor placement for fault diagnosis.
\newblock {\em IEEE Transactions on Systems, Man, and Cybernetics-Part A:
  Systems and Humans}, 38(6):1398--1410, 2008.

\bibitem{wu2020comprehensive}
Zonghan Wu, Shirui Pan, Fengwen Chen, Guodong Long, Chengqi Zhang, and S~Yu
  Philip.
\newblock A comprehensive survey on graph neural networks.
\newblock {\em IEEE transactions on neural networks and learning systems},
  32(1):4--24, 2020.

\bibitem{bronstein2021geometric}
Michael~M Bronstein, Joan Bruna, Taco Cohen, and Petar Veli{\v{c}}kovi{\'c}.
\newblock Geometric deep learning: Grids, groups, graphs, geodesics, and
  gauges.
\newblock {\em arXiv preprint arXiv:2104.13478}, 2021.

\bibitem{scarselli2008graph}
Franco Scarselli, Marco Gori, Ah~Chung Tsoi, Markus Hagenbuchner, and Gabriele
  Monfardini.
\newblock The graph neural network model.
\newblock {\em IEEE transactions on neural networks}, 20(1):61--80, 2008.

\bibitem{bruna2013spectral}
Joan Bruna, Wojciech Zaremba, Arthur Szlam, and Yann LeCun.
\newblock Spectral networks and locally connected networks on graphs.
\newblock {\em arXiv preprint arXiv:1312.6203}, 2013.

\bibitem{Kipf2016-vy}
Thomas~N Kipf and Max Welling.
\newblock Semi-supervised classification with graph convolutional networks.
\newblock {\em arXiv preprint arXiv:1609.02907}, 2016.

\bibitem{duvenaud2015convolutional}
David~K Duvenaud, Dougal Maclaurin, Jorge Iparraguirre, Rafael Bombarell,
  Timothy Hirzel, Al{\'a}n Aspuru-Guzik, and Ryan~P Adams.
\newblock Convolutional networks on graphs for learning molecular fingerprints.
\newblock {\em Advances in neural information processing systems}, 28, 2015.

\bibitem{Yu2017-td}
Bing Yu, Haoteng Yin, and Zhanxing Zhu.
\newblock Spatio-temporal graph convolutional networks: A deep learning
  framework for traffic forecasting.
\newblock {\em arXiv preprint arXiv:1709.04875}, 2017.

\bibitem{Chen2019-md}
Weiqi Chen, Ling Chen, Yu~Xie, Wei Cao, Yusong Gao, and Xiaojie Feng.
\newblock Multi-range attentive bicomponent graph convolutional network for
  traffic forecasting.
\newblock In {\em Proceedings of the AAAI conference on artificial
  intelligence}, volume~34, pages 3529--3536, 2020.

\bibitem{Wu2022-af}
Shiwen Wu, Fei Sun, Wentao Zhang, Xu~Xie, and Bin Cui.
\newblock Graph neural networks in recommender systems: A survey.
\newblock {\em ACM Comput. Surv.}, 55(5):1--37, December 2022.

\bibitem{Schlichtkrull2018-ae}
Michael Schlichtkrull, Thomas~N Kipf, Peter Bloem, Rianne van~den Berg, Ivan
  Titov, and Max Welling.
\newblock Modeling relational data with graph convolutional networks.
\newblock In {\em The Semantic Web}, pages 593--607. Springer International
  Publishing, 2018.

\bibitem{zhu2021survey}
Yanqiao Zhu, Weizhi Xu, Jinghao Zhang, Yuanqi Du, Jieyu Zhang, Qiang Liu, Carl
  Yang, and Shu Wu.
\newblock A survey on graph structure learning: Progress and opportunities.
\newblock {\em arXiv preprint arXiv:2103.03036}, 2021.

\bibitem{franceschi2019learning}
Luca Franceschi, Mathias Niepert, Massimiliano Pontil, and Xiao He.
\newblock Learning discrete structures for graph neural networks.
\newblock In {\em International conference on machine learning}, pages
  1972--1982. PMLR, 2019.

\bibitem{yu2021graph}
Donghan Yu, Ruohong Zhang, Zhengbao Jiang, Yuexin Wu, and Yiming Yang.
\newblock Graph-revised convolutional network.
\newblock In {\em Machine Learning and Knowledge Discovery in Databases:
  European Conference, ECML PKDD 2020, Ghent, Belgium, September 14--18, 2020,
  Proceedings, Part III}, pages 378--393. Springer, 2021.

\bibitem{jin2020graph}
Wei Jin, Yao Ma, Xiaorui Liu, Xianfeng Tang, Suhang Wang, and Jiliang Tang.
\newblock Graph structure learning for robust graph neural networks.
\newblock In {\em Proceedings of the 26th ACM SIGKDD international conference
  on knowledge discovery \& data mining}, pages 66--74, 2020.

\bibitem{wang2021graph}
Ruijia Wang, Shuai Mou, Xiao Wang, Wanpeng Xiao, Qi~Ju, Chuan Shi, and Xing
  Xie.
\newblock Graph structure estimation neural networks.
\newblock In {\em Proceedings of the Web Conference 2021}, pages 342--353,
  2021.

\bibitem{chen2020iterative}
Yu~Chen, Lingfei Wu, and Mohammed Zaki.
\newblock Iterative deep graph learning for graph neural networks: Better and
  robust node embeddings.
\newblock {\em Advances in neural information processing systems},
  33:19314--19326, 2020.

\bibitem{Fatemi2021-vy}
Bahare Fatemi, Layla El~Asri, and Seyed~Mehran Kazemi.
\newblock Slaps: Self-supervision improves structure learning for graph neural
  networks.
\newblock {\em Advances in Neural Information Processing Systems},
  34:22667--22681, 2021.

\bibitem{rozemberczki2021pytorch}
Benedek Rozemberczki, Paul Scherer, Yixuan He, George Panagopoulos, Alexander
  Riedel, Maria Astefanoaei, Oliver Kiss, Ferenc Beres, Guzmán López, Nicolas
  Collignon, and Rik Sarkar.
\newblock Pytorch geometric temporal: Spatiotemporal signal processing with
  neural machine learning models, 2021.

\bibitem{zhu2020beyond}
Jiong Zhu, Yujun Yan, Lingxiao Zhao, Mark Heimann, Leman Akoglu, and Danai
  Koutra.
\newblock Beyond homophily in graph neural networks: Current limitations and
  effective designs.
\newblock {\em Advances in Neural Information Processing Systems},
  33:7793--7804, 2020.

\bibitem{Bai2018-zs}
Shaojie Bai, J~Zico Kolter, and Vladlen Koltun.
\newblock An empirical evaluation of generic convolutional and recurrent
  networks for sequence modeling.
\newblock {\em arXiv preprint arXiv:1803.01271}, 2018.

\bibitem{Xu2018-hs}
Keyulu Xu, Weihua Hu, Jure Leskovec, and Stefanie Jegelka.
\newblock How powerful are graph neural networks?
\newblock {\em arXiv preprint arXiv:1810.00826}, 2018.

\bibitem{Kuramoto1975-yk}
Yoshiki Kuramoto.
\newblock Self-entrainment of a population of coupled non-linear oscillators.
\newblock In {\em International Symposium on Mathematical Problems in
  Theoretical Physics}, pages 420--422. Springer Berlin Heidelberg, 1975.

\bibitem{paszke2019pytorch}
Adam Paszke, Sam Gross, Francisco Massa, Adam Lerer, James Bradbury, Gregory
  Chanan, Trevor Killeen, Zeming Lin, Natalia Gimelshein, Luca Antiga, et~al.
\newblock Pytorch: An imperative style, high-performance deep learning library.
\newblock {\em Advances in neural information processing systems}, 32, 2019.

\bibitem{kuramoto}
Fabrizio Damicelli.
\newblock Python implementation of the kuramoto model.
\newblock \url{https://github.com/fabridamicelli/kuramoto}, 2019.

\bibitem{kingma2014adam}
Diederik~P Kingma and Jimmy Ba.
\newblock Adam: A method for stochastic optimization.
\newblock {\em arXiv preprint arXiv:1412.6980}, 2014.

\end{thebibliography}

\end{document}